\documentclass[runningheads]{llncs}
\usepackage{graphicx}
\usepackage{comment}
\usepackage{amsmath,amssymb} %
\usepackage{color}

\usepackage{algorithm}
\usepackage{algpseudocode}
\usepackage{makecell}
\usepackage{booktabs}
\usepackage{epsfig}
\usepackage{subfigure}
\usepackage{mydefs}
\usepackage{microtype}
\usepackage{tablefootnote}
\def\segmethodfirst{SemSeg}
\def\segmethodsecond{SiamSeg}
\def\segmethodthird{FewShotSeg}

\def\bboracle{\bbox\text{-oracle}}

\begin{document}
\pagestyle{headings}
\mainmatter
\def\ECCVSubNumber{3}  %

\title{An Exploration of \\ Target-Conditioned Segmentation Methods \\ for Visual Object Trackers} %

\titlerunning{Exploring Target-Dependent Segmentation Methods for Trackers}
\author{Matteo Dunnhofer\orcidID{0000-0002-1672-667X}, 
Niki Martinel\orcidID{0000-0002-6962-8643}, \\ and
Christian Micheloni\orcidID{0000-0003-4503-7483}}
\authorrunning{M. Dunnhofer et al.}
\institute{Machine Learning and Perception Lab, University of Udine, Italy}
\maketitle

\begin{abstract}
Visual object tracking is the problem of predicting a target object's state in a video. Generally, bounding-boxes have been used to represent states, and a surge of effort has been spent by the community to produce efficient causal algorithms capable of locating targets with such representations. As the field is moving towards binary segmentation masks to define objects more precisely, in this paper we propose to extensively explore target-conditioned segmentation methods available in the computer vision community, in order to transform any bounding-box tracker into a segmentation tracker. Our analysis shows that such methods allow trackers to compete with recently proposed segmentation trackers, while performing quasi real-time.
\keywords{Visual Object Tracking, Video Object Segmentation, Target-Conditioned Segmentation, Deep Learning.}
\end{abstract}

\section{Introduction}
In its simplest definition, visual object tracking corresponds to the %
prediction of the state of a target object in a stream of images. It is considered one of the most difficult problems in computer vision. 
Object occlusion and fast motion, light changes, and motion blur are some of the challenges that algorithms have to deal with.
Additionally, constraints of real-time operation are often demanded by the many applications, such as video surveillance, behavior understanding, autonomous driving, and robotics.

Despite the many state representation one can use to model the target's state, the bounding-box has been the most used until now. This is a rectangle that encloses the object of interest, and it is defined by the coordinates of its top-left corner or of its center, and by its width and height. Based on this representation, many model-free tracking algorithms have been studied.
Early solutions were based on mean shift algorithms \cite{Comanciu2000}, part-based methods \cite{LGT,OGT}, and SVM learning \cite{Struck}.
Later, the correlation filter approach gained interest thanks to its fast processing time \cite{Bolme2010,KCF,DSST,Staple,Lukezic2018}. 
More recently, the features of convolutional neural networks (CNNs) have been exploited to develop more efficient trackers. Trackers based on this image representation include deep regression networks \cite{GOTURN,RE3,Dunnhofer2020distilled}, online tracking-by-detection methods \cite{MDNet,RealTimeMDNet}, solutions that use reinforcement learning \cite{Yun2017,Ren2018,Chen2018,Dunnhofer2019,Dunnhofer2020distilled}, CNN-based discriminative correlation filters \cite{CCOT,ECO,ATOM,DiMP} and siamese CNNs \cite{SiamFC,SiamRPN,SiamRPNpp,DaSiam,Zhang2019,SiamMask}. The last two methods raised the state-of-the-art year-by-year, showing a remarkable performance across all the available tracking benchmarks \cite{GOT10k,LaSOT,OTB,UAV123,TrackingNet,VOT2018,VOT2019} that almost reaches the 70\% of bounding-box overlap accuracy.
Such a high and generalized performance poses the question about 
if the bounding-box based measures have been now saturated. 
Moreover, it was proven in the object detection community that humans hardly distinguish a bounding-box prediction that has a 30\% overlap from one with 50\% \cite{Russakovsky2015}. Hence, it is fair to ask ourselves if a 100\% overlap accuracy tracker is really necessary for applications. From such considerations, we can wonder if the time is done for bounding-box representations. 

Furthermore, starting from 2020, tracking communities (VOT2020\footnote{\code{\url{https://votchallenge.net/vot2020/}}} and MOT\footnote{\code{\url{https://motchallenge.net}}}) raised the bar in their annual challenges by requesting trackers binary segmentation maps -- precise location and shape definitions through pixel-wise background-target classifications -- as target state representation.
Segmentation representations are not new in the visual tracking panorama. In many applications, model-based algorithms used contours \cite{Yang2005,Li2005} or masks \cite{McFarlane1995,Kim2002,Dunnhofer2020MedIA} for tracking particular objects. From a more general point of view, the recent video object segmentation (VOS) problem requires to produce the segmentation masks of generic target objects in a video, given the mask of each in the first frame. The currently available solutions propose highly accurate methods in terms of segmentation ability \cite{OSVOS,RGMP,OnAVOS,PReMVOS,OSMN,BoLTVOS}, but with the drawback of poor speed performance. This is due to characteristics of the available benchmarks \cite{DAVIS2016,DAVIS2017,YouTubeVOS}, that do not include challenging situations from a tracking point of view. 
In fact, these datasets provide temporally short sequences where the target covers a large fraction of frames, its appearance does not suffer major changes, or low background distractors are present. The performance of such methods on standard VOT benchmarks was proven very poor \cite{VOT2018} and to mitigate such behavior, the SiamMask \cite{SiamMask} and D3S \cite{D3S} algorithms have been proposed recently. These solutions adapted, respectively, the siamese correlation approach and discriminative correlation filters to segmentation outputs, and showed promising results while performing in real-time.

We believe that the huge effort spent by the tracking community in developing bounding-box based trackers can be still exploited in the  segmentation tracking domain.
With such an idea in mind, in this paper, we propose to explore what is currently available in the computer vision literature that can be adapted to make any bounding-box tracker output segmentation masks. In particular, we propose to extensively evaluate three methods: Box2Seg \cite{BoLTVOS}, SiamMask \cite{SiamMask}, and AMP \cite{AMP}. Two were already proposed for this task \cite{BoLTVOS,AMP}, but their capabilities were little explored. The other is a recent segmentation tracker \cite{SiamMask} that we reinterpret as a segmentation module.
Our evaluations are based on a framework that requires a bounding-box tracker to provide a coarse localization of the object, and then a segmentation module conditioned on the target object is employed to provide its precise localization. 
Along with practical considerations, we will show that this combination can produce trackers able to compete with the recently proposed methods \cite{SiamMask,D3S} on the VOT2020 and DAVIS  benchmarks.

\subsection{Related Work}
Combining segmentation methods and trackers has been increasingly tackled in the last two years. SiamMask \cite{SiamMask} and D3S \cite{D3S} employed a CNN decoder module \cite{Ronneberger2015,Pinheiro2016} to refine a latent representation constructed by a cross-correlation operation and discriminative filter, respectively. %
Zhang \etal \cite{Zhang2018} proposed to use ECO tracker's \cite{ECO} bounding-box predictions to improve the segmentation performance of the OSVOS \cite{OSVOS} VOS method. Similarly, \cite{BoLTVOS} adapted a deep CNN for semantic segmentation to generate a segmentation mask after the bounding-box proposal of a tracking-by-detection approach. The combination of these methods achieved promising results, but they were mainly focused on the VOS task. Additionally, they did not provide any extensive evaluation by considering different trackers and segmentation methods. In this paper, we aim to provide a deep analysis of such combination on both visual tracking and VOS benchmarks.

\section{Methodology}
\label{sec:segmodels}
In this paper, we study how state-of-the-art off-the-shelf bounding-box trackers can be augmented to track an object with the requirement of a segmentation representation. Our idea is based on the belief that the much effort spent in developing algorithms to predict the motion of a target is relevant even if a segmentation is required. 
To implement our analysis, we design a framework where a bounding-box tracker is first used to get a coarse localization of the target object, and then a target-conditioned segmentation method is executed to generate a pixel-wise map. Under this setup, any bounding-box based tracker can be transformed into a segmentation tracker.
Considering separately tracking and mask generation carries practical advantages: (i) the performance of a segmentation tracker can be analyzed more consistently, by separating the error committed in the localization from the error in shape definition; (ii) flexibility of easily switch tracking and segmentation modules to adapt to application needs; (iii) availability of two different forms of output (bounding-box and mask) that are obtained with independent modules.

In the following of this section, we first introduce the framework employed for the analysis. Then, an abstract description of each of the selected segmentation methods will be given.

\begin{figure}[!t]%
\begin{center}
\includegraphics[width=.8\columnwidth]{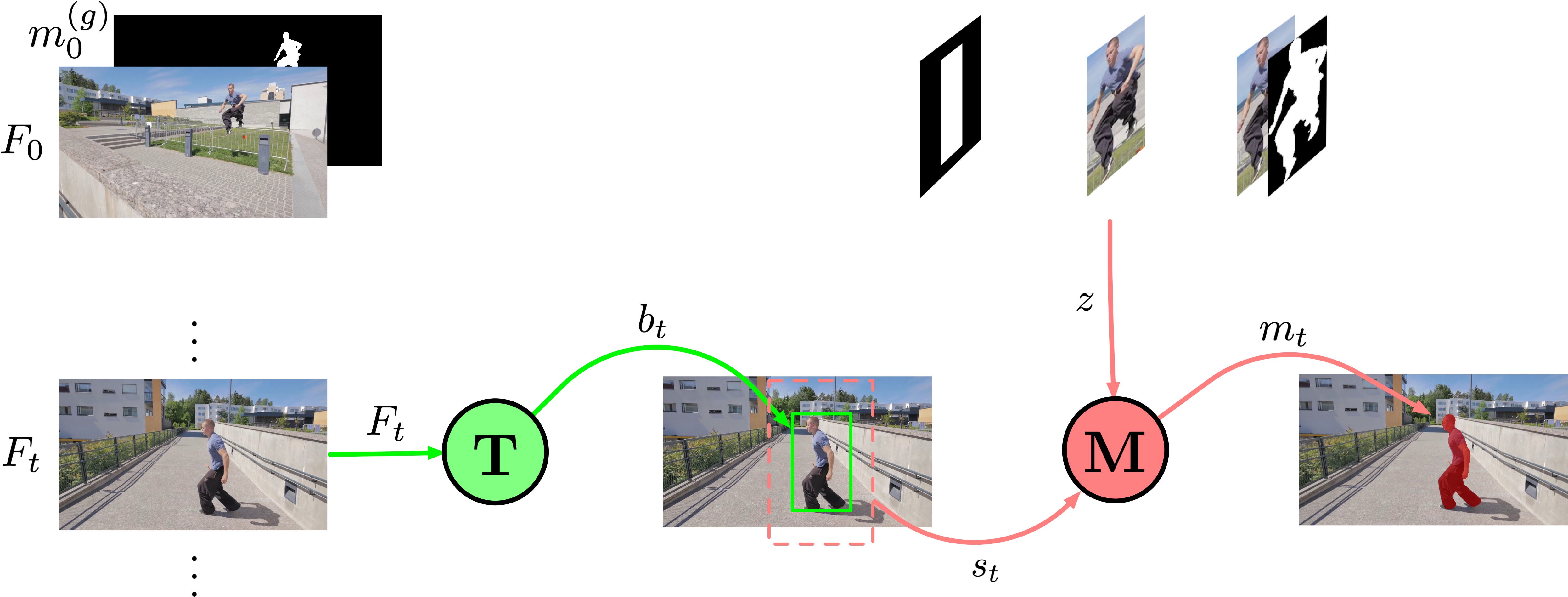}
\end{center}
   \caption{Graphical representation of the framework used for the evaluation. $\maskgt_0$ outlines the target to be tracked in the first frame $\frame_0$ of a video. At every step $t$, the frame $\frame_t$ is first given in input the the tracker $\tracker$ which outputs a bounding-box estimate $\bbox_t$. This, together with a factor $k$, is employed to crop a searching area $\sa_t$ in $\frame_t$. $\sa_t$ is inputted to the mask generation algorithm $\mg$ which is conditioned on the target template $\template$ computed in different form depending on $\mg$'s input requirements. $\mg$ returns the segmentation of the target inside $\sa_t$. The output mask $\mask_t$ is finally built by placing $\mg$'s output inside a zero-matrix at the location of $\sa_t$. }
\label{fig:framework}
\end{figure}

\subsection{Segmentation Tracking Framework}
We first define the key elements of the framework.
A video 
\begin{align}
    \video = \{ \frame_t \}, t \in \{0, \cdots, T \}, T \in \integers
\end{align} 
is considered as a $T$ long sequence of frames $\frame_t \in \images$, where $\images = \{0, \cdots, 255\}^{W \times H \times 3}$ is the space of RGB images. 
We treat a bounding-box tracking algorithm as a function
\begin{align}
    \tracker: \images \rightarrow \reals^4
\end{align}
that is inputted with frame $\frame_t$ and produces a bounding-box estimate $\bbox_t = [x_t, y_t, w_t, h_t]$ as a real-valued vector containing the center coordinates $x_t, y_t$, and the width and height $w_t$, $h_t$ (in the image coordinate system).\footnote{At $t=0$, $\tracker$ is initialized with $\frame_0$ and the ground-truth bounding-box $\bboxgt_0$.}
In a similar fashion, we consider a target-based segmentation algorithm as the function
\begin{align}
    \mg: \patches \times \templates \rightarrow \{ 0, 1 \}^{W' \times H'}
\end{align}
which is given an image patch $\sa_t \in \patches = \{0, \cdots, 255\}^{W' \times H' \times 3} \subseteq \images$ extracted from $F_t$ and a template image $\template \in \templates$ of the target object, and outputs a binary segmentation mask with zero-elements belonging to the background and one-elements defining the pixels of the target.

Given these concepts, the segmentation tracking procedure works as follows. At every time step $t$ of a video $\video$, $\frame_t$ is first given to the tracker $\tracker$ to produce $\bbox_t$. Then, $\frame_t$ and $\bbox_t$ are used to extract a searching area 
\begin{align}
    \sa_t = F_t[x_t, y_t, k \cdot w_t, k \cdot h_t]
\end{align} 
which is the area of $\frame_t$ localized by the coordinates of $\bbox_t$ and which width and height are scaled by the factor $k \in \reals$. $\sa_t$ and $\template$ are given to the segmentation algorithm $\mg$ to produce the pixel-wise mask of the target inside $\sa_t$.\footnote{The details to obtain $\template$ are described in each subsection describing the segmentation methods.} 
The output mask $\mask_t$ is finally built by placing $\mg$'s output at the $\sa_t$ location of a zero-matrix with size $W \times H$. 

A graphical representation of the described framework is shown in Figure \ref{fig:framework}.

\subsection{Target-Conditioned Segmentation Methods}
In this subsection we describe the target-conditioned segmentation methodologies we analyzed. Three conceptually different approaches were chosen:
\begin{itemize}
    \item an adapted semantic segmentation network \cite{BoLTVOS,DeepLabv3}, that we name \segmethodfirst;
    \item a module based on the siamese correlation framework \cite{SiamMask}, referred as \segmethodsecond;
    \item a few-shot segmentation algorithm \cite{AMP}, called \segmethodthird.
\end{itemize}

\begin{figure}[!t]%
\begin{center}
\includegraphics[width=.9\columnwidth]{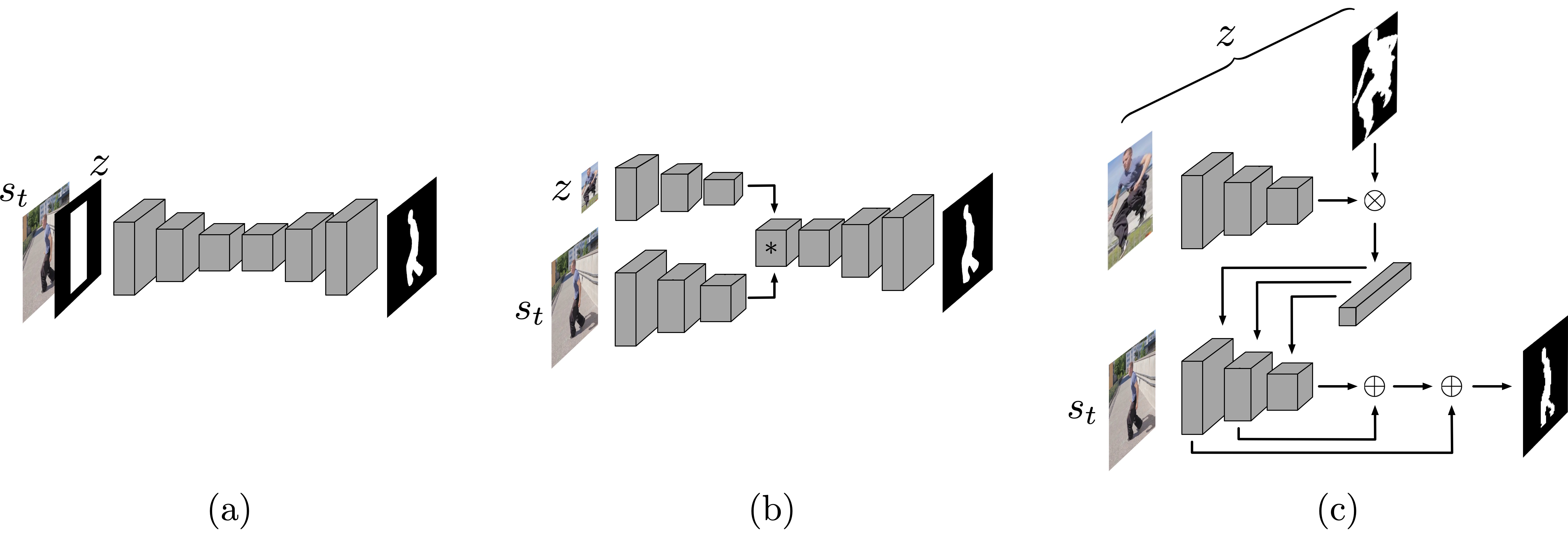}
\end{center}
   \caption{Visual representation of the methodologies employed to generate target segmentations. (a) shows \segmethodfirst, a deep segmentation model adapted to take in input a 4-channel tensor composed of $\sa_t$ and $\template$ and produce the mask of the object. (b) presents \segmethodsecond, the siamese framework where a cross correlation operation between $\template$ and $\sa_t$ features is employed to first locate the object, and then to produce its segmentation mask. (c) shows \segmethodthird, a few-shot segmentation algorithm that is adapted for visual segmentation tracking, by considering $\template$ as the support set and $\sa_t$ as the query image.}
\label{fig:segmethods}
\end{figure}

\subsubsection{\segmethodfirst.}
\label{sec:segdeep}
The first target-conditioned segmentation method we analyzed was proposed as Box2Seg in \cite{PReMVOS,BoLTVOS}. The idea is to adapt a state-of-the-art fully convolutional deep neural network for image segmentation to target segmentation. Given an RGB image and an additional input channel containing coarse information regarding the position of the target, this module produces a detailed segmentation of the latter. 
In the context of our framework, the RGB channels of the searching area $\sa_t$ are concatenated with the template channel
\begin{align}
    \template = \{0, 1\}^{k\cdot w_t, k \cdot h_t}
\end{align} 
which is a binary mask of the same size of $\sa_t$, and which positive elements are located inside the area defined by $\bbox_t$. $\template$ is computed at every time step $t$, and the 4-channel input resulting from the concatenation is given to the network which produces the segmentation of the target inside the searching area. A visualization of this approach is proposed in Figure \ref{fig:segmethods} (a). 
The network is trained offline by exploiting object segmentation, instance segmentation, and/or VOS datasets. The training pairs are formed as batches of inputs-targets, where the first are composed by searching area and template (built using the bounding-box that encloses the ground-truth segmentation), and the second are the actual object masks. Optimization is done by solving a two class segmentation problem (foreground-background) defined as the minimization of a pixel-wise classification loss (cross-entropy, Dice loss, etc.).
This approach has the advantage of requiring just the bounding-box as first-frame target definition.

\subsubsection{\segmethodsecond.}
\label{sec:segsiam}
As second mask generation method, we reinterpreted the siamese correlation framework for segmentation tracking \cite{SiamMask}.
The general view of this scheme is to first locate the target template in the higher-level feature space of template and searching area, and then project the localization into the segmentation space. These steps are jointly implemented with an encoder-decoder CNN architecture, which %
capabilities are acquired through an end-to-end offline procedure in which the whole model is optimized by minimizing a foreground-background pixel-wise classification loss. The training examples are pairs of searching area-template inputs, and ground-truth target masks, where searching area and template are sampled without temporal correlation.

Following this intuition, we adapted such method in our framework as follows.  The target template is the image crop
\begin{align}
    z = F_0[x_0, y_0, w_0, h_0]
\end{align}
extracted from the first frame $\frame_0$ of the video, using the ground-truth bounding-box $\bboxgt_0 = [x_0, y_0, w_0, h_0]$. $\bboxgt_0$ is obtained as the box that encloses the ground-truth mask $\maskgt_0$. The features $\widehat{\template}$ of $\template$ are computed with a forward pass through the encoder module just at $t = 0$. At every other $t$, $\widehat{\template}$ is cross-correlated to the encoded representation $\widehat{\sa_t}$, and the resulting activation map is then refined by the decoder module, and ultimately placed into $\mask_t$. 
The procedure is depicted in Figure \ref{fig:segmethods} (b) and, as for \segmethodfirst, it just requires the target definition as a bounding-box.

\subsubsection{\segmethodthird.}
\label{sec:segfew}
The last analyzed methodology treats target-conditioned segmentation as a few-shot segmentation problem. In such a setting, the goal is to provide a pixel-wise segmentation of a target object inside a query image, given a so-called support-set, i.e. one or more (few-shot) image and mask examples of the target. Algorithms for this problem are generally designed as fully CNNs, where the segmentation ability is guided by other convolutional branches or by model parameters that are made dependent on the support-set.

This view of few-shot segmentation can be reframed for the purpose of segmentation tracking. In our setting, the support-set is considered as the target template 
\begin{align}
    z = (F_0[x_0, y_0, w_0, h_0], \maskgt_0[x_0, y_0, w_0, h_0] )
\end{align}
that is the pair of the image crop that contains the visual appearance of the target in $\frame_0$, and the relative cropped ground-truth mask. %
The crops are constructed considering $\bboxgt_0 = [x_0, y_0, w_0, h_0]$.
The searching area $\sa_t$ is extracted after every $\bbox_t$ of $\tracker$ and it is considered as the query image. Together with the template (the support-set), they are given to the few-shot segmentation model to produce the target segmentation. A graphical example of this methodology is proposed in Figure \ref{fig:segmethods} (c).
With respect to the previous methods, employing \segmethodthird\ requires the definition of the target object through a mask.

\section{Experimental Setup}
In this section, we report the experimental procedures we performed to implement and analyze the previously presented methodologies. All experiments were run on a machine with an Intel Xeon E5-2690 v4 @ 2.60GHz CPU, 320 GB of RAM, and an NVIDIA TITAN V GPU. Code for tracker and segmentation methods was implemented in Python.

\subsection{Trackers}
The trackers selected for the analysis were KCF \cite{KCF}, DCFNet \cite{DCFNet}, MDNet \cite{MDNet}, MetaCrest \cite{MetaTrackers}, SiamFC \cite{SiamFC}, SiamRPN \cite{SiamRPN}, ECO \cite{ECO}, ATOM \cite{ATOM}, and DiMP \cite{DiMP}.
Such algorithms were chosen because they tackle visual tracking by different approaches and so can provide performance of various quality. 
For each of them, we used the public code made available by the authors. We tried the best to respect default parameters and settings.

\subsection{Segmentation Modules}

\subsubsection{\segmethodfirst.} 
To implement this methodology, we followed the details of the Box2Seg refinement module provided in \cite{BoLTVOS,PReMVOS}. The DeepLab-v3 architecture \cite{DeepLabv3} for image segmentation was translated for the task of interest. ImageNet \cite{ImageNet} pre-trained ResNet-50 \cite{He2016ResNet} was employed as backbone network and adapted to receive the 4-channel tensor. Before being inputted, the concatenated RGB and template channels were resized to $385 \times 385$ pixels. During training, the searching area was enlarged by the factor $k$, chosen uniformly in $\{ 1, 1.25, 1.5, 1.75, 2\}$. Batches of 12 input-target mask pairs were sampled from a training set composed of the training sets of COCO \cite{COCO}, YouTube-VOS \cite{YouTubeVOS}, and DAVIS 2016 and 2017 \cite{DAVIS2016,DAVIS2017}. Learning rate was set to $10^{-5}$ for the backbone layers, and to $10^{-4}$ for all the others. Training was carried on until the mIoU \cite{PascalVOC}, computed over the foreground and background classes, stopped improving on a custom validation set composed of the validation sets of the aforementioned datasets.

\subsubsection{\segmethodsecond.}
The second approach introduced in subsection \ref{sec:segsiam} was implemented through the segmentation tracker SiamMask \cite{SiamMask}. We used the code provided by the authors along with the pre-trained models. For completion, we present to the reader some information about the training procedures performed by the authors. The SiamMask architecture model was trained in two-stages: first, the encoder module based on ResNet-50 \cite{He2016ResNet} was trained for target localization by optimizing a multi-task loss for similarity maximization and RPN \cite{FasterRCNN} detection. After that, the decoder module designed as \cite{Pinheiro2016} was attached to the intermediate cross-correlation map and trained by minimizing a foreground-background pixel-wise cross-entropy loss.
The training set used was a combination of ImageNet-VID \cite{ImageNet}, COCO \cite{COCO} and YouTube-VOS \cite{YouTubeVOS}.
Before being inputted to the model, $\template$ and $\sa_t$ were resized to $127 \times 127$ and $255 \times 255$ pixels respectively.

\subsubsection{\segmethodthird.}
As a few-shot segmentation module, we employed the strategy proposed in \cite{AMP}, which is a recently introduced state-of-the-art method that has been shown to perform well also in VOS tasks. 
The authors proposed a sample efficient method to segment an unseen class object via a multi-resolution imprinting procedure of adaptive masked proxies (AMP). AMPs are constructed by a Normalized Masked Average Pooling (NMAP) operation between the CNN embeddings of the support set's RGB sample and its relative binary mask. The AMP representations are used to imprint \cite{Qi2018}, at multiple resolutions, the CNN embeddings computed on the query image. The VGG16 \cite{VGG} architecture is employed as a backbone feature extractor, and skip connections are also exploited as done similarly in FCN8s \cite{Shelhamer2017}. Data extracted from the PASCAL-VOC dataset \cite{PascalVOC} was used to compose training samples as query image, support-set image, support-set mask, and target mask. Optimization was performed by minimizing the pixel-wise cross-entropy loss between predicted and ground-truth masks.
Code and pre-trained model provided by the authors were adapted to our implementation needs.

\subsection{Benchmarks and Performance Measures}
We performed analysis on the VOT2020 benchmark, and the validation sets of the DAVIS 2016 \cite{DAVIS2016} and DAVIS 2017 \cite{DAVIS2017} VOS benchmarks. All provide segmentations as target representations.

For VOT2020 we employed the newly introduced protocol.\footnote{\code{\url{https://data.votchallenge.net/vot2020/vot-2020-protocol.pdf}}} The novel baseline protocol requires running a tracker on shorter sequences determined by predefined points (anchors). From such starting points, the tracker is initialized with the ground-truth mask and run either forward or backward, depending on the longest sub-sequence yielded by the two directions. %
The new accuracy ($\vota$) measures the average pixel-wise intersection-over-union between predicted and ground-truth masks, for frames where the tracker did not fail (i.e. the accuracy did not decrease after a certain threshold). The new robustness ($\votr$) expresses the normalized average number of frames where the algorithm successfully tracked the target before drifting.
The two measures are joined in a refreshed single performance score known as expected average overlap ($\voteao$). 
Version 0.4.2 of the Python toolkit was used to obtain the results.

The protocol used for DAVIS datasets is similar to the One-Pass evaluation (OPE) employed in OTB \cite{OTB} benchmarks: the tracker is initialized with the mask of the target object in the first frame, and then it is run until the end of the sequence. Performance is measured in terms of the Jaccard index $\davisj$ which measures the pixel-wise intersection-over-union between the predicted and ground-truth masks. Along with this index, the F-measure $\davisf$ is employed to evaluate contour accuracy. For both measures, mean ($\davisjm, \davisfm$), recall ($\davisjr, \davisfr$), and decay ($\davisjd, \davisfd$) values are reported. For DAVIS 2017, where multiple objects must be tracked and segmented, we run the trackers independently for each object and then fuse the prediction masks by assigning each pixel to the object that received higher confidence in that location.

\begin{table*}[t]
	\fontsize{5}{6}\selectfont
	\centering
	\caption{Results of the baseline experiment on VOT2020. Best segmentation method results, per tracker, are highlighted in red (Rectangular Mask results are excluded).}
	\label{tab:vot2020}
	\begin{tabulary}{1\linewidth}{L{=5.5em} | *{3}{^C{3.35em}} | *{3}{^C{3.35em}} | *{3}{^C{3.35em}} | *{3}{^C{3.35em}} }
		\toprule
					 & \multicolumn{3}{c|}{\segmethodfirst} & \multicolumn{3}{c|}{\segmethodsecond}  & \multicolumn{3}{c|}{\segmethodthird}  & \multicolumn{3}{c}{Rectangular Mask} \\
		\multirow{-2}{*}{Tracker} & $\voteao$ & $\vota$ & $\votr$ & $\voteao$ & $\vota$ & $\votr$ & $\voteao$ & $\vota$ & $\votr$ & $\voteao$ & $\vota$ & $\votr$ \\
		\midrule
		DCFNet  & 0.203 & 0.616 & 0.426 & \tblbest{0.310} & \tblbest{0.676} & 0.558 & 0.230 & 0.491 & \tblbest{0.567} & 0.184 & 0.441 & 0.523 \\
		KCF  & 0.199 & 0.648 & 0.371 & \tblbest{0.285} & \tblbest{0.659} & \tblbest{0.501} & 0.200 & 0.459 & 0.499 & 0.155 & 0.402 & 0.432 \\
		SiamFC  & 0.218 & 0.602 & 0.446 & \tblbest{0.309} & \tblbest{0.682} & \tblbest{0.571} & 0.228 & 0.491 & 0.563 & 0.183 & 0.418 & 0.537 \\
		MetaCrest  & 0.240 & 0.602 & 0.513 & \tblbest{0.336} & \tblbest{0.657} & 0.624 & 0.250 & 0.479 & \tblbest{0.647} & 0.189 & 0.390 & 0.587 \\
		SiamRPN  & 0.356 & 0.692 & 0.639 & \tblbest{0.369} & \tblbest{0.701} & 0.651 & 0.311 & 0.551 & \tblbest{0.677} & 0.247 & 0.452 & 0.663 \\
		MDNet  & 0.295 & 0.638 & 0.609 & \tblbest{0.371} & \tblbest{0.662} & 0.689 & 0.308 & 0.546 & \tblbest{0.723} & 0.234 & 0.440 & 0.687 \\
		ATOM  & 0.402 & 0.678 & \tblbest{0.735} & \tblbest{0.406} & \tblbest{0.691} & 0.723 & 0.337 & 0.560 & 0.731 & 0.277 & 0.467 & 0.738 \\
		DiMP  & 0.410 & 0.675 & 0.744 & \tblbest{0.410} & \tblbest{0.691} & 0.730 & 0.347 & 0.556 & \tblbest{0.749} & 0.278 & 0.464 & 0.733 \\
		ECO  & 0.322 & 0.632 & 0.735 & \tblbest{0.414} & \tblbest{0.694} & 0.729 & 0.349 & 0.561 & \tblbest{0.759} & 0.275 & 0.459 & 0.746 \\
		
		\midrule
		
		$\bboracle$ & \tblbest{0.806} & \tblbest{0.809} & \tblbest{0.996} & 0.697 & 0.744 & 0.970 & 0.541 & 0.623 & 0.941 & 0.516 & 0.519 & 1.0  \\
		\bottomrule		
\end{tabulary}
\end{table*}

\section{Results}

\subsubsection{General Performance.} 
Results on VOT2020 benchmark are presented in Table \ref{tab:vot2020}. Trackers combined with \segmethodsecond\ achieve the best overall performance in $\voteao$ and $\vota$. This is explained by the fact the VOT benchmarks include difficult tracking scenarios for trackers, resulting in lower quality bounding-boxes that affect \segmethodfirst\ and \segmethodthird.  %
Thanks to its more robust segmentation method, \segmethodsecond\ allows to recover (to some extent) from inaccurate $\bbox_t$ estimates and so produce more accurate target segmentations. Interestingly, \segmethodthird\ is the approach that achieves the highest $\votr$, showing to be the method less susceptible to failure. For all the methods, employing a better tracker is fundamental to improve the overall performance.

\begin{table*}[t]
	\fontsize{5}{6}\selectfont
	\centering
	\caption{$\davisj$ results on DAVIS 2016 validation set. Best segmentation method results, per tracker, are highlighted in red (Rectangular Mask results are excluded).}
	\label{tab:davis2016j}
	\begin{tabulary}{1\linewidth}{L{=6em} | *{3}{^C{3.2em}} | *{3}{^C{3.2em}} | *{3}{^C{3.2em}} | *{3}{^C{3.2em}} }
		\toprule
					 & \multicolumn{3}{c|}{\segmethodfirst} & \multicolumn{3}{c|}{\segmethodsecond}  & \multicolumn{3}{c|}{\segmethodthird}  & \multicolumn{3}{c}{Rectangular Mask} \\
		\multirow{-2}{*}{Tracker} & $\davisjm$ & $\davisjr$ & $\davisjd$ & $\davisjm$ & $\davisjr$ & $\davisjd$ & $\davisjm$ & $\davisjr$ & $\davisjd$ & $\davisjm$ & $\davisjr$ & $\davisjd$ \\
		\midrule
		KCF  & 0.527 & 0.570 & 0.174 & 0.557 & 0.616 & 0.199 & \tblbest{0.580} & \tblbest{0.688} & \tblbest{0.162} & 0.302 & 0.200 & 0.153 \\
		DCFNet  & 0.531 & 0.574 & 0.209 & 0.551 & 0.627 & 0.229 & \tblbest{0.564} & \tblbest{0.674} & \tblbest{0.178} & 0.313 & 0.183 & 0.130 \\
		MetaCrest  & 0.574 & 0.624 & 0.169 & 0.595 & 0.672 & 0.145 & \tblbest{0.598} & \tblbest{0.712} & \tblbest{0.136} & 0.323 & 0.151 & 0.108 \\
		MDNet  & 0.582 & 0.635 & 0.177 & 0.593 & 0.656 & 0.196 & \tblbest{0.610} & \tblbest{0.717} & \tblbest{0.143} & 0.342 & 0.198 & 0.149 \\
		SiamFC  & 0.607 & 0.661 & \tblbest{0.159} & 0.611 & 0.694 & 0.177 & \tblbest{0.621} & \tblbest{0.738} & 0.163 & 0.356 & 0.234 & 0.140 \\
		ECO  & 0.615 & 0.679 & \tblbest{0.099} & 0.623 & 0.744 & 0.108 & \tblbest{0.626} & \tblbest{0.748} & 0.113 & 0.375 & 0.243 & 0.070 \\
		SiamRPN  & \tblbest{0.689} & 0.772 & \tblbest{0.089} & 0.663 & 0.782 & 0.111 & 0.681 & \tblbest{0.859} & \tblbest{0.089} & 0.417 & 0.340 & 0.066 \\
		ATOM  & \tblbest{0.723} & \tblbest{0.846} & \tblbest{0.074} & 0.658 & 0.785 & 0.105 & 0.669 & 0.845 & 0.081 & 0.415 & 0.345 & 0.053 \\
		DiMP  & \tblbest{0.723} & 0.827 & \tblbest{0.086} & 0.704 & 0.844 & 0.100 & 0.699 & \tblbest{0.886} & 0.095 & 0.443 & 0.379 & 0.027 \\
		
		\midrule
		
		$\bboracle$ & \tblbest{0.812} & 0.920 & \tblbest{0.020} & 0.732 & 0.896 & 0.044 & 0.739 & \tblbest{0.946} & 0.052 & 0.455 & 0.418 & 0.008 \\
		\bottomrule		
\end{tabulary}
\end{table*}
\begin{table*}[t]
	\fontsize{5}{6}\selectfont
	\centering
	\caption{$\davisf$ results on DAVIS 2016 validation set. Best segmentation method results, per tracker, are highlighted in red (Rectangular Mask results are excluded).}
	\label{tab:davis2016f}
	\begin{tabulary}{1\linewidth}{L{=6em} | *{3}{^C{3.2em}} | *{3}{^C{3.2em}} | *{3}{^C{3.2em}} | *{3}{^C{3.2em}} }
		\toprule
					 & \multicolumn{3}{c|}{\segmethodfirst} & \multicolumn{3}{c|}{\segmethodsecond}  & \multicolumn{3}{c|}{\segmethodthird}  & \multicolumn{3}{c}{Rectangular Mask} \\
		\multirow{-2}{*}{Tracker} & $\davisfm$ & $\davisfr$ & $\davisfd$ & $\davisfm$ & $\davisfr$ & $\davisfd$  & $\davisfm$ & $\davisfr$ & $\davisfd$ & $\davisfm$ & $\davisfr$ & $\davisfd$  \\
		\midrule
		DCFNet  & \tblbest{0.553} & 0.587 & 0.210 & 0.536 & 0.596 & 0.187 & 0.530 & \tblbest{0.599} & \tblbest{0.158} & 0.155 & 0.017 & 0.068 \\
		KCF  & \tblbest{0.559} & 0.577 & 0.180 & 0.525 & 0.572 & 0.190 & 0.542 & \tblbest{0.598} & \tblbest{0.149} & 0.136 & 0.018 & 0.119 \\
		MetaCrest  & \tblbest{0.599} & 0.632 & 0.193 & 0.561 & \tblbest{0.637} & \tblbest{0.136} & 0.572 & 0.634 & 0.137 & 0.139 & 0.019 & 0.063 \\
		MDNet  & \tblbest{0.603} & 0.623 & \tblbest{0.170} & 0.570 & 0.616 & 0.197 & 0.582 & \tblbest{0.635} & \tblbest{0.170} & 0.163 & 0.050 & 0.112 \\
		SiamFC  & \tblbest{0.633} & 0.665 & \tblbest{0.152} & 0.592 & 0.663 & 0.159 & 0.597 & \tblbest{0.675} & 0.157 & 0.156 & 0.037 & 0.126 \\
		ECO  & \tblbest{0.637} & \tblbest{0.696} & \tblbest{0.097} & 0.590 & 0.692 & 0.102 & 0.592 & 0.673 & 0.117 & 0.170 & 0.020 & 0.066 \\
		SiamRPN  & \tblbest{0.713} & \tblbest{0.783} & \tblbest{0.105} & 0.629 & 0.707 & 0.127 & 0.642 & 0.752 & \tblbest{0.105} & 0.186 & 0.059 & 0.081 \\
		ATOM  & \tblbest{0.739} & \tblbest{0.856} & 0.098 & 0.628 & 0.697 & 0.111 & 0.626 & 0.751 & \tblbest{0.090} & 0.178 & 0.025 & 0.060 \\
		DiMP  & \tblbest{0.744} & \tblbest{0.821} & \tblbest{0.108} & 0.658 & 0.754 & 0.130 & 0.657 & 0.767 & 0.118 & 0.191 & 0.071 & 0.015 \\
		
		\midrule
		
		$\bboracle$ & \tblbest{0.843} & \tblbest{0.918} & \tblbest{0.033} & 0.693 & 0.805 & 0.064 & 0.717 & 0.873 & 0.056 & 0.219 & 0.073 & 0.015 \\
		\bottomrule		
\end{tabulary}
\end{table*}

Results on the DAVIS 2016 benchmark are reported in Tables \ref{tab:davis2016j} and \ref{tab:davis2016f}. More weak trackers like DCFNet, KCF, MDNet, MetaCrest, and SiamFC, benefit of \segmethodthird\ for pixel-wise accuracy. When more precise bounding-box estimates are provided, through ECO, SiamRPN, ATOM, DiMP, \segmethodfirst\ allows the best $\davisjm$ performance. For $\davisjr$ and $\davisjd$, \segmethodthird\ is almost always the best approach.
For contour accuracy, \segmethodfirst\ is generally the best method at $\davisfm$. Better trackers also benefit the same for $\davisfr$ and $\davisfd$. For the others, \segmethodthird\ gets the best results.
\segmethodsecond\ is the weakest method on this benchmark, justified by the presence of easy tracking situations that put the focus on providing more accurate target segmentations.

\begin{table*}[t]
	\fontsize{5}{6}\selectfont
	\centering
	\caption{$\davisj$ results on DAVIS 2017 validation set. Best segmentation method results, per tracker, are highlighted in red (Rectangular Mask results are excluded).}
	\label{tab:davis2017j}
	\begin{tabulary}{1\linewidth}{L{=6em} | *{3}{^C{3.2em}} | *{3}{^C{3.2em}} | *{3}{^C{3.2em}} | *{3}{^C{3.2em}} }
		\toprule
					 & \multicolumn{3}{c|}{\segmethodfirst} & \multicolumn{3}{c|}{\segmethodsecond}  & \multicolumn{3}{c|}{\segmethodthird}  & \multicolumn{3}{c}{Rectangular Mask} \\
		\multirow{-2}{*}{Tracker} & $\davisjm$ & $\davisjr$ & $\davisjd$ & $\davisjm$ & $\davisjr$ & $\davisjd$ & $\davisjm$ & $\davisjr$ & $\davisjd$ & $\davisjm$ & $\davisjr$ & $\davisjd$ \\
		\midrule
		DCFNet  & 0.443 & 0.474 & 0.299 & \tblbest{0.455} & \tblbest{0.497} & 0.281 & 0.424 & 0.434 & \tblbest{0.214} & 0.283 & 0.166 & 0.176 \\
		KCF  & 0.433 & 0.464 & 0.277 & \tblbest{0.461} & \tblbest{0.517} & 0.272 & 0.425 & 0.451 & \tblbest{0.209} & 0.268 & 0.167 & 0.198 \\
		MDNet  & 0.444 & 0.478 & 0.284 & \tblbest{0.465} & \tblbest{0.515} & 0.260 & 0.444 & 0.493 & \tblbest{0.216} & 0.284 & 0.156 & 0.168 \\
		MetaCrest  & 0.447 & 0.468 & 0.276 & \tblbest{0.468} & \tblbest{0.518} & 0.262 & 0.426 & 0.443 & \tblbest{0.178} & 0.273 & 0.145 & 0.155 \\
		SiamFC  & 0.466 & 0.499 & 0.260 & \tblbest{0.468} & \tblbest{0.523} & 0.277 & 0.431 & 0.454 & \tblbest{0.225} & 0.280 & 0.176 & 0.196 \\
		ECO  & \tblbest{0.498} & 0.556 & 0.244 & 0.503 & \tblbest{0.567} & 0.222 & 0.458 & 0.501 & \tblbest{0.178} & 0.310 & 0.220 & 0.132 \\
		SiamRPN  & \tblbest{0.536} & \tblbest{0.600} & 0.233 & 0.506 & 0.578 & 0.237 & 0.470 & 0.518 & \tblbest{0.180} & 0.321 & 0.248 & 0.141 \\
		ATOM  & \tblbest{0.566} & \tblbest{0.659} & \tblbest{0.148} & 0.544 & 0.626 & 0.188 & 0.488 & 0.547 & 0.168 & 0.321 & 0.251 & 0.103 \\
		DiMP  & \tblbest{0.583} & \tblbest{0.671} & \tblbest{0.148} & 0.553 & 0.639 & 0.170 & 0.498 & 0.555 & 0.162 & 0.323 & 0.251 & 0.093 \\
		
		\midrule
		
		$\bboracle$ & \tblbest{0.762} & \tblbest{0.891} & \tblbest{0.0} & 0.618 & 0.738 & 0.073 & 0.578 & 0.694 & 0.059 & 0.408 & 0.340 & 0.0 \\
		\bottomrule		
\end{tabulary}
\end{table*}
\begin{table*}[t]
	\fontsize{5}{6}\selectfont
	\centering
	\caption{$\davisf$ results on DAVIS 2017 validation set. Best segmentation method results, per tracker, are highlighted in red (Rectangular Mask results are excluded).}
	\label{tab:davis2017f}
	\begin{tabulary}{1\linewidth}{L{=6em} | *{3}{^C{3.2em}} | *{3}{^C{3.2em}} | *{3}{^C{3.2em}} | *{3}{^C{3.2em}} }
		\toprule
					 & \multicolumn{3}{c|}{\segmethodfirst} & \multicolumn{3}{c|}{\segmethodsecond}  & \multicolumn{3}{c|}{\segmethodthird}  & \multicolumn{3}{c}{Rectangular Mask} \\
		\multirow{-2}{*}{Tracker} & $\davisfm$ & $\davisfr$ & $\davisfd$ & $\davisfm$ & $\davisfr$ & $\davisfd$  & $\davisfm$ & $\davisfr$ & $\davisfd$ & $\davisfm$ & $\davisfr$ & $\davisfd$ \\
		\midrule
		KCF  & \tblbest{0.517} & 0.542 & 0.307 & 0.500 & 0.544 & 0.287 & 0.506 & \tblbest{0.565} & \tblbest{0.266} & 0.172 & 0.035 & 0.178 \\
		DCFNet  & \tblbest{0.532} & 0.567 & 0.322 & 0.512 & 0.561 & 0.284 & 0.511 & \tblbest{0.572} & \tblbest{0.237} & 0.194 & 0.049 & 0.140 \\
		MDNet  & 0.525 & 0.563 & 0.288 & 0.513 & 0.565 & 0.270 & \tblbest{0.526} & \tblbest{0.598} & \tblbest{0.255} & 0.184 & 0.059 & 0.170 \\
		MetaCrest  & \tblbest{0.545} & \tblbest{0.593} & 0.305 & 0.520 & 0.567 & 0.278 & 0.521 & 0.583 & \tblbest{0.241} & 0.176 & 0.042 & 0.146 \\
		SiamFC  & \tblbest{0.556} & \tblbest{0.611} & 0.299 & 0.523 & 0.583 & 0.294 & 0.524 & 0.601 & \tblbest{0.267} & 0.184 & 0.064 & 0.184 \\
		ECO  & \tblbest{0.592} & \tblbest{0.663} & 0.255 & 0.553 & 0.620 & 0.236 & 0.553 & 0.637 & \tblbest{0.226} & 0.214 & 0.055 & 0.128 \\
		SiamRPN  & \tblbest{0.626} & \tblbest{0.713} & 0.259 & 0.552 & 0.628 & 0.257 & 0.567 & 0.670 & \tblbest{0.219} & 0.210 & 0.070 & 0.150 \\
		DiMP  & \tblbest{0.663} & \tblbest{0.765} & \tblbest{0.181} & 0.591 & 0.675 & 0.215 & 0.584 & 0.685 & 0.195 & 0.206 & 0.059 & 0.093 \\
		ATOM  & \tblbest{0.640} & \tblbest{0.751} & \tblbest{0.195} & 0.584 & 0.664 & 0.218 & 0.574 & 0.666 & 0.201 & 0.203 & 0.053 & 0.104 \\
		
		\midrule
		
		$\bboracle$ & \tblbest{0.829} & \tblbest{0.945} & \tblbest{0.017} & 0.654 & 0.779 & 0.097 & 0.685 & 0.847 & 0.078 & 0.280 & 0.116 & 0.028 \\
		\bottomrule		
\end{tabulary}
\end{table*}

On DAVIS 2017, which results are presented in Tables \ref{tab:davis2017j} and \ref{tab:davis2017f}, \segmethodfirst\ is still the best approach to use with stronger bounding-box trackers for $\davisjm$ and $\davisjr$. \segmethodthird\ is the method that achieves the most consistent masks across time. For low-performance tracking algorithms, \segmethodsecond\ results to be better than the others in $\davisjm$ and $\davisjr$, mitigating the lower tracking performance with its target search strategy and showing the increased difficulty of this benchmark than its previous version.
In terms of contour performance, \segmethodfirst\ is the most appropriate method for $\davisfm$ and $\davisfr$ performance. For $\davisfd$, \segmethodthird\ results in the best solution.
Overall, as for VOT2020, in both DAVIS 2016 and 2017 employing better trackers lets achieve the best performances.

\subsubsection{Comparison with a Rectangular Segmentation Tracker.} In the last block of columns of Tables \ref{tab:vot2020}, \ref{tab:davis2016j}, \ref{tab:davis2016f}, \ref{tab:davis2017j}, \ref{tab:davis2017f}, we report the performance of the trackers considering their $\bbox_t$ predictions as  segmentation masks, i.e. binary mask where the rectangular area defined by $\bbox_t$ is filled with 1. 
Overall, all the considered segmentation methods improve those baseline results on all the benchmarks and across all measures. This proves that employing the approaches presented in this paper lets bounding-box trackers improve their accuracy in terms of precise target definition.
\segmethodfirst\ is the method that achieves generally the best improvement, followed by \segmethodsecond\ and \segmethodthird.

\subsubsection{Comparison with a Bounding-box Oracle Tracker.} In the last row of Tables \ref{tab:vot2020}, \ref{tab:davis2016j}, \ref{tab:davis2016f}, \ref{tab:davis2017j}, \ref{tab:davis2017f}, the performance of a bounding-box oracle based tracker, $\bboracle$ (i.e. the tracker that returns the ground-truth bounding-box $\bboxgt_t$ at every $t$), is presented. 
Given this ground-truth information, \segmethodfirst\ is the approach that best segments the target object, on every considered benchmark and performance measure. On VOT2020, accuracy and robustness performances reach almost 80\% and 100\%, meaning that its segmentation capabilities are effective for the objects contained in this dataset. \segmethodsecond\ follows with a decrease of 9\% and 2.6\%, while \segmethodthird\ shows a much bigger performance loss in $\vota$  (-25\% than \segmethodfirst) than in $\votr$ (-5.5\%).
\segmethodthird\ comes after \segmethodfirst\ in terms of $\davisj$ and $\davisf$ on DAVIS 2016, and in terms of $\davisf$ on DAVIS 2017. 
\segmethodsecond\ gets the weakest performance on DAVIS 2016 but surpasses \segmethodthird\ in $\davisj$ on DAVIS 2017.

\segmethodfirst\ is also the method that suffers the major gap between the $\bboracle$ performance and the best tracker DiMP ($\voteao$ loss -48\%, average $\davisjm$ loss -17.2\%, average $\davisfm$ loss -15.9\%). This shows the susceptibility to misaligned bounding-box predictions (we hypothesize this can be mitigated introducing some noise to the input bounding-boxes in the training procedure). The performance decrease happens also for the other methods, although with less magnitude.

\subsubsection{Separating Localization and Segmentation Error.} The results obtained with $\bboracle$ and the rectangular mask output allow us to determine the tracking and segmentation error committed by $\tracker$ and $\mg$ respectively. 
The error $e_{\tracker}$ committed by the tracker is just the performance difference between $\bboracle$ and $\tracker$, both considered with rectangular mask output. The error $e_{\mg}$ of $\mg$ can be computed as the performance difference between $\bboracle$ with $\mg$ and $\tracker$ with $\mg$ which tracking performance is corrected by summing $e_{\tracker}$. In this setting, $e_{\mg}$ is considered as the distance from $\mg$'s maximum achievable performance, that happens when $\bboracle$ is employed as tracker. For example, when MDNet and \segmethodfirst\ are executed together, the $\vota$ error $e_{\tracker}$ is computed as $e_{\tracker} = 0.519 - 0.440 = 0.079$, while the $e_{\mg}$ is obtained as $e_{\mg} = 0.809 - (0.638 + 0.079) = 0.092$. So, it results that the highest loss in accuracy is due to the segmentation than to tracking. If DiMP and \segmethodsecond\ are considered, we have an $\vota$ error $e_{\tracker} = 0.055$ and $e_{\tracker} = -0.002$ meaning that \segmethodsecond\ compensates the tracking error and even improves the performance of the combination.

\begin{figure}[!t]%
\begin{center}
\includegraphics[width=.8\columnwidth]{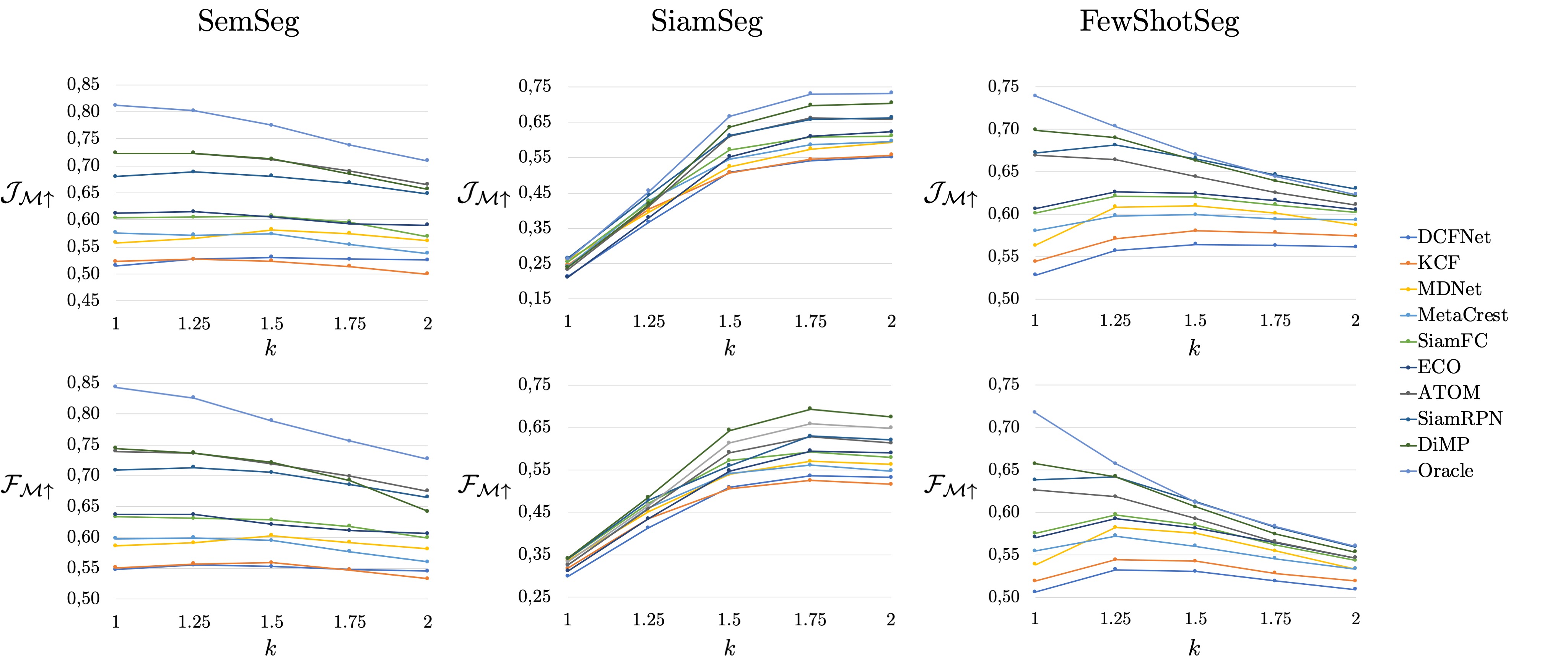}
\end{center}
   \caption{Results on the DAVIS 2016 validation set of the sensibility of the target segmentation methods to the size of the searching area. Performance is evaluated in terms of $\davisjm$ and $\davisfm$.}
\label{fig:sadavis2016}
\end{figure}

\subsubsection{Impact of the Searching Area Size.}
We analyzed how sensible the three segmentation methods are to different sizes of the searching area. In particular, the factor $k$ was studied across the values \{1, 1.25, 1.5, 1.75, 2\} on the DAVIS 2016 benchmark, and the results are shown in Figure \ref{fig:sadavis2016}.
With \segmethodfirst, all the trackers show a slow decrease in $\davisjm$ and $\davisfm$ performance by enlarging the searching area. Best performance are obtained with $k = 1$ or $k = 1.25$ (proven also by the $\bboracle$ based tracker). 
Similar conclusions can be made for \segmethodthird. The highest $\davisjm$ is achieved with $k = 1.25$. For larger $k$, the performance of more weak trackers remains constant, while the performance of stronger trackers slightly decreases. $\davisfm$ tends to decrease for all the trackers.
\segmethodsecond\ shows the opposite trend. Better results are obtained with larger searching areas. Specifically, best $\davisjm$ and $\davisfm$ performance are obtained with $k \geq 1.75$. Weaker trackers have a smaller performance decrease between 1.75 and 1.5 than stronger ones, while for $k < 1.5$ the performance of all the trackers quickly drops. This can be explained by \segmethodsecond's training methodology, where the objective is set as target localization and segmentation in large image patches.

\begin{table*}[t]
	\fontsize{5}{6}\selectfont
	\centering
	\caption{Results on the speed analysis (in seconds and FPS) of the combined tracker-segmentation methods. The original tracker speeds are reported in the last two columns (times of the employed implementations). The last row shows the average speed of running just the segmentation methods.}
	\label{tab:speed}
	\begin{tabulary}{1\linewidth}{L{=6em} | *{2}{^C{3.2em}} | *{2}{^C{3.2em}} | *{2}{^C{3.2em}} | *{2}{^C{3.2em}} }
		\toprule
					 & \multicolumn{2}{c|}{\segmethodfirst} & \multicolumn{2}{c|}{\segmethodsecond}  & \multicolumn{2}{c|}{\segmethodthird}  & \multicolumn{2}{c}{Tracker speed} \\
		\multirow{-2}{*}{Tracker} & s & FPS & s & FPS & s & FPS & s & FPS \\
		\midrule
		MDNet  & 0.628 & 1.6 & 0.580 & 1.7 & 0.704 & 1.4 & 0.550 & 1.8  \\
		MetaCrest  & 0.181 & 5.5 & 0.134 & 7.5 & 0.258 & 3.9 & 0.109 & 9.1  \\
		ECO  & 0.126 & 8 & 0.081 & 12.4 & 0.220 & 4.6 & 0.059 & 17.0  \\
		ATOM  & 0.123 & 8.1 & 0.079 & 12.7 & 0.198 & 5.1 & 0.050 & 20.0  \\
		DiMP  & 0.109 & 9.1 & 0.068 & 14.7 & 0.189 & 5.3 & 0.038 & 26.3  \\
		SiamRPN  & 0.026 & 12.1 & 0.047 & 21.4 & 0.172 & 5.8 & 0.026 & 38.4  \\
		KCF  & 0.070 & 14.3 & 0.034 & 29.5 & 0.167 & 6.0 & 0.013 & 78.8  \\
		SiamFC  & 0.064 & 15.6 & 0.030 & 33.5 & 0.157 & 6.4 & 0.008 & 125.3  \\
		DCFNet  & 0.062 & 16.2 & 0.026 & 39.0 & 0.143 & 7.0 & 0.004 & 227.8 \\

		\midrule
		
		No tracker & 0.062 & 16.3 & 0.024 & 42.8 & 0.151 & 6.7 & - & -   \\
		\bottomrule		
\end{tabulary}
\end{table*}
\subsubsection{Speed Analysis.} In Table \ref{tab:speed} an analysis of the speed of the algorithms is presented. The fastest method to produce a segmentation is \segmethodsecond\ which runs at 43 FPS. With this method, DCFNet and SiamFC run in real-time (39 and 34 FPS respectively). Stronger trackers like ECO, ATOM, and DiMP, achieve a speed of 12, 13, and 15 FPS respectively.
\segmethodfirst\ runs independently at 16 FPS, and combined with SiamRPN and DiMP allows a speed of 12 and 9 FPS respectively.
\segmethodthird\ is the slowest method and takes around 7 FPS. In this setup, the speed performance is almost completely taken by the segmentation method and best trackers reach a speed of 5-6 FPS.

\subsubsection{State-of-the-art Comparison.} Comparison with the state-of-the-art is presented in Table \ref{tab:sota}. The VOS methods outperform every studied $\tracker-\mg$ combination on DAVIS 2016 and 2017, but they show poor speed results. DiMP and ATOM with \segmethodfirst\ perform better than SiamMask in $\davisj$ on both DAVIS 2016 and 2017. In terms of $\davisf$ they outperform also D3S. On DAVIS 2017, D3S is improved by DiMP-\segmethodfirst\ in every measure. On VOT2020, SiamMask is largely beaten by all the best trackers, combined both with \segmethodfirst\ and \segmethodsecond. All the trackers using the second method improves SiamMask, showing its limitations in target localization.
ECO and \segmethodsecond\ reaches an $\voteao$ of 0.414, slightly improving DiMP and ATOM. With the same segmentation method, SiamRPN outperforms D3S in $\vota$, achieving the best 0.701, while maintaining a quasi real-time speed of 21 FPS. 

\begin{table*}[t]
	\fontsize{5}{5.5}\selectfont
	\centering
	\caption{State-of-the-art comparison for the best combinations. Best results are highlighted in red, second-best in blue.}%
	\label{tab:sota}
	\resizebox{\textwidth}{!}{
	\begin{tabular}{l | c c c c | c c c c | c c c | c c c }
		\toprule
					 & \multicolumn{4}{c|}{DAVIS 2016} & \multicolumn{4}{c|}{DAVIS 2017}  & \multicolumn{3}{c|}{VOT2020} &  \\
		\multirow{-2}{*}{Method} & $\davisjm$ & $\davisjr$ & $\davisfm$ & $\davisfr$ & $\davisjm$ & $\davisjr$ & $\davisfm$ & $\davisfr$ & $\voteao$ & $\vota$ & $\votr$  & \multirow{-2}{*}{FPS}\\
		\midrule
		OSMN \cite{OSMN} & 0.740 & 0.876 & 0.729 & 0.840 & 0.525 & 0.609 & 0.571 & 0.661 & - & - & - & 7 \\
		BoLTVOS \cite{BoLTVOS} & 0.781 & - & 0.812 & - & \tblsecondbest{0.684} & - & \tblsecondbest{0.754} & - & - & - & - & 1 \\
		OSVOS \cite{OSVOS} & 0.798 & 0.936 & 0.806 & \tblsecondbest{0.926} & 0.566 & \tblsecondbest{0.638} & 0.639 & \tblsecondbest{0.738} & - & - & - & 0.1 \\
		FAVOS \cite{FAVOS} & 0.824 & \tblbest{0.965} & 0.795 & 0.894 & 0.546 & 0.611 & 0.618 & 0.723 & - & - & - & 0.8 \\
		RGMP \cite{RGMP} & 0.815 & 0.917 & 0.820 & 0.908 & 0.648 & - & 0.686 & - & - & - & - & 8 \\
		OnAVOS \cite{OnAVOS} & \tblbest{0.857} & - & \tblsecondbest{0.842} & - & 0.610 & - & 0.661 & - & - & - & - & 0.1 \\
		PReMVOS \cite{PReMVOS} & \tblsecondbest{0.849} & \tblsecondbest{0.961} & \tblbest{0.886} & \tblbest{0.947} & \tblbest{0.739} & \tblbest{0.831} & \tblbest{0.817} & \tblbest{0.889} & - & - & - & 0.03 \\
		\midrule
		
		SiamMask\footnotemark  & 0.692 & 0.848 & 0.639 & 0.743 & 0.522 & 0.597 & 0.559 & 0.645 & 0.321 & 0.686 & 0.569  & 43 \\
		D3S \cite{D3S} & 0.754 & - & 0.726 & - & 0.578 & - & 0.638 & - & \tblbest{0.439} & \tblsecondbest{0.699} & \tblbest{0.769}  & 25 \\
		
		\midrule
		SiamRPN-\segmethodsecond  & 0.663 & 0.782 & 0.629 & 0.707 & 0.506 & 0.578 & 0.552 & 0.628 & 0.369 & \tblbest{0.701} & 0.651 & 21  \\
		ECO-\segmethodsecond  & 0.623 & 0.744 & 0.590 & 0.692 & 0.503 & 0.567 & 0.553 & 0.620 & \tblsecondbest{0.414} & 0.694 & 0.729 & 12 \\
		ATOM-\segmethodsecond  & 0.658 & 0.785 & 0.628 & 0.697 & 0.544 & 0.626 & 0.584 & 0.664 & 0.406 & 0.691 & 0.723 & 13  \\
		ATOM-\segmethodfirst  & 0.723 & 0.846 & 0.739 & 0.856 & 0.566 & 0.659 & 0.640 & 0.751 & 0.402 & 0.678 & 0.735 & 8  \\
		DiMP-\segmethodfirst  & 0.723 & 0.827 & 0.744 & 0.821 & 0.583 & 0.671 & 0.663 & 0.765 & 0.410 & 0.675 & \tblsecondbest{0.744}  & 9 \\
		DiMP-\segmethodsecond  & 0.704 & 0.844 & 0.658 & 0.754 & 0.553 & 0.639 & 0.591 & 0.675 & 0.410 & 0.691 & 0.730 & 15  \\
		\bottomrule		
\end{tabular}
}
\end{table*}

\footnotetext{Since we used SiamMask to implement \segmethodsecond, for fair comparison we report the results of the same implementation used for segmentation tracking, which has slightly worse performance than presented in the original paper.}

\begin{figure}[!t]%
\begin{center}
\includegraphics[width=.9\columnwidth]{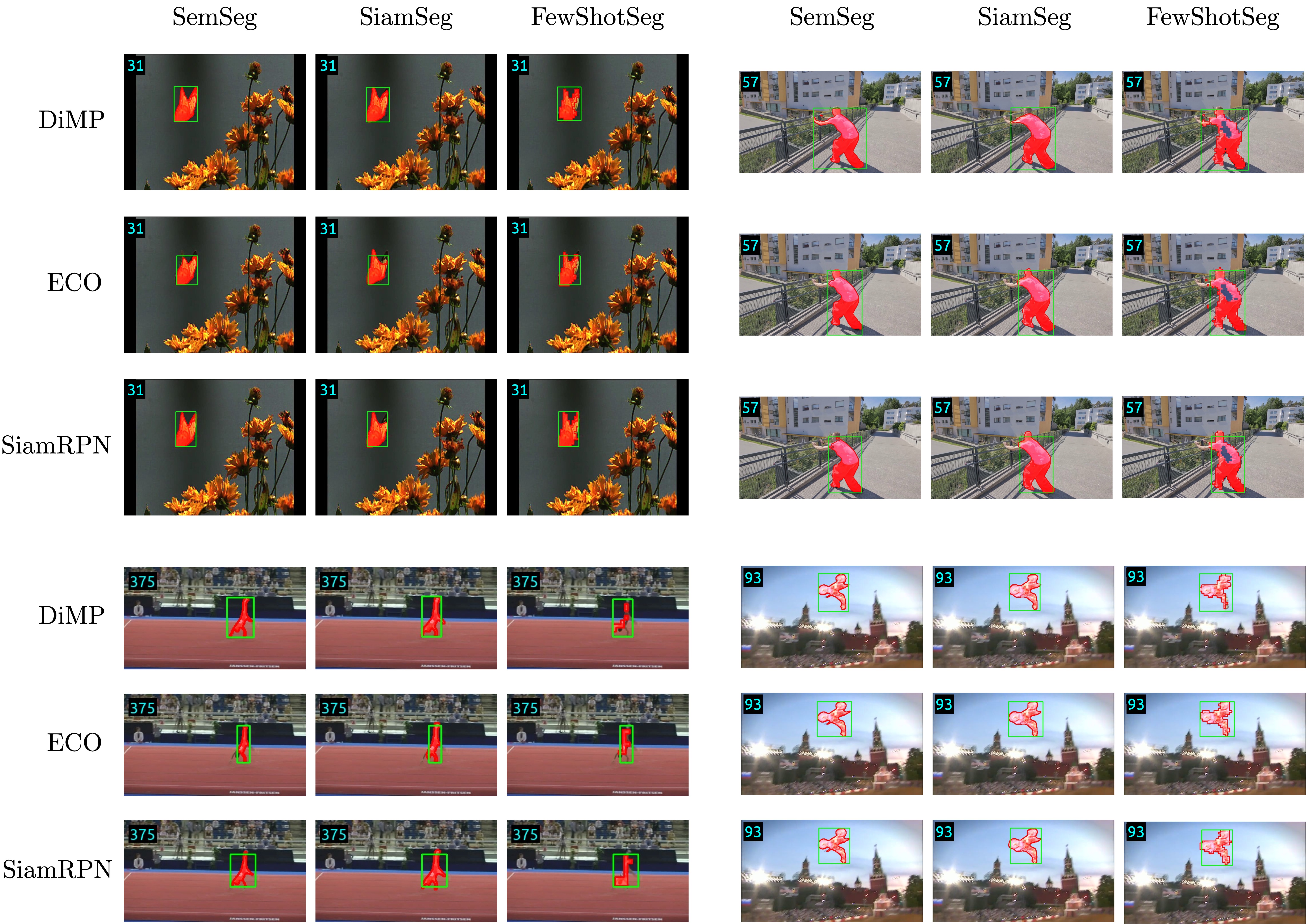}
\end{center}
   \caption{Qualitative examples of the segmentation (red superimposed mask) proposed by the three target-conditioned segmentation methods, based on the bounding-box proposals (green rectangles) given by three different trackers.}
\label{fig:qualitativeex}
\end{figure}

In Figure \ref{fig:qualitativeex} some qualitative examples of the segmentation methods are proposed.\footnote{For more, please see \href{https://youtu.be/SODiKBD84_g}{\texttt{ https://youtu.be/SODiKBD84\_g}}.}

\section{Conclusions}
In this paper, three target-conditioned segmentation methods, \segmethodfirst, \segmethodsecond, and \segmethodthird, were extensively analyzed to transform any bounding-box tracker into a segmentation tracker. \segmethodfirst\ and \segmethodsecond\ resulted in the stronger methods, and their combination with trackers like SiamRPN, ECO, ATOM, and DiMP, allows to compete with the most recent segmentation trackers SiamMask and D3S on the DAVIS 2016 and 2017, and VOT2020 benchmarks.

\subsubsection{Acknowledgements.} This work was supported by the ACHIEVE-ITN H2020 project.

\bibliographystyle{splncs04}
\bibliography{egbib}

\clearpage

\end{document}